\documentclass[conference]{IEEEtran}
\usepackage{cite}
\usepackage{hyperref}

\ifCLASSINFOpdf
  
\else
\fi

\usepackage{graphicx}

\begin{document}

\title{Shape Representation and Classification through Pattern Spectrum and Local Binary Pattern - A Decision Level Fusion Approach}


\author{\IEEEauthorblockN{B.H.Shekar}
\IEEEauthorblockA{Department of Computer Science,\\ Mangalore University,\\Mangalore, Karnataka,\\ India.\\
Email: bhshekar@gmail.com}
%\and
%\IEEEauthorblockN{R.K.Bharathi}
%\IEEEauthorblockA{Dept. of Master of Computer Applications,\\ 
%S.J.College of Engineering,\\Mysore, Karnataka,\\India.\\
%Email: rkbharathi@hotmail.com}
\and
\IEEEauthorblockN{Bharathi Pilar}
\IEEEauthorblockA{Dept. of Master of Computer Applications,\\ 
AIMIT, St. Aloysius College, \\Mangalore, Karnataka, \\India.\\
Email: bharathi.pilar@gmail.com}
}

%\author{\IEEEauthorblockN{Bharathi Pilar}
%\IEEEauthorblockA{Department of Master of Computer Applications\\
%AIMIT, St. Aloysius college,Mangalore, Karnataka,India.\\
%Email: bharathi.pilar@gmail.com}


\maketitle

\begin{abstract}
%\boldmath
In this paper, we present a decision level fused local Morphological Pattern Spectrum(PS) and Local Binary Pattern (LBP) approach for an efficient shape representation and classification. This method makes use of Earth Movers Distance(EMD) as the measure in feature matching and shape retrieval process.  The proposed approach has three major phases : Feature Extraction,  Construction of hybrid spectrum knowledge base and Classification. In the first phase, feature extraction of the shape is done using pattern spectrum and  local binary pattern method. In the second phase, the histograms of both pattern spectrum and local binary pattern are fused and stored in the knowledge base. In the third phase, the comparison and matching of the features, which are represented in the form of histograms, is done using Earth Movers Distance(EMD) as metric. The top-n shapes are retrieved for each query shape. The accuracy is tested by means of standard Bulls eye  score method. The experiments are conducted on  publicly available shape datasets like Kimia-99, Kimia-216 and  MPEG-7. The comparative study is also provided with the well known approaches to exhibit the retrieval accuracy of the proposed approach. 
\end{abstract}

% \fi
\IEEEpeerreviewmaketitle

\section{Introduction}
In the field of computer vision, the process of understanding the scene involves object extraction, representation or feature extraction, matching and classification. In particular, object representation and classification constitute a deeply entrenched and ubiquitous component of  any automated intelligent systems. In order to classify the object, representing the object in suitable form with dominant features is very much essential. It is observed from the literature that the shape is one such dominant feature for object representation. A good shape representation should be compact and should retain the essential characteristics of the shape. It should be invariant to rotation and scale. In this paper, the representation process takes shape of the object as feature since shape is an important,effective and rich feature which is used to describe both image boundary as well as content and hence plays a significant role in classification task. 

Shape representation and description techniques can be generally classified into two broad categories: Contour-based methods and Region-based methods. Contour based shape techniques exploit shape boundary information, where as, in region based techniques, all the pixels within a shape region are taken into account to obtain the shape representation. Contour-based approaches are more popular than region-based approaches in literature since human beings are thought to discriminate shapes mainly by their contour features. However, region-based methods are more robust as they use the entire shape information. These methods can cope well with shape defection which arises due to missing shape part or occlusion. Skeleton or medial axis transformation is an example for region based technique. In our work, both contour as well as skeletons are used to take advantage of both the methods. We present a novel approach devised based on the morphological pattern spectrum to obtain pattern spectrum for every binary shape making use of their contour and skeleton. Then the local binary pattern(LBP) of binary shape is computed. The Earth Movers Distance(EMD) measure is used to compare and match the pattern spectrum of the training set and the query shape.  

 The remaining part of the paper is organized as follows. In section 2, the review of the related works are brought down. In section 3, the technique of morphological patter spectrum is given. In section 4,  an insight into local binary pattern is presented, followed by the description about the EMD metric in section 5. The proposed approach is given in section 6. In section 7, the  experimental set-up along with the discussion of the results are brought out  and conclusion is given in section 8.

\section{Review of Related work}
The general principle adopted for shape representation and classification involves two basic steps: representation of an object retaining the dominant features and matching the features with a suitable distance measure. We have seen several methods in the literature, which have been developed for shape representation and classification. More in particular from the field of mathematical morphology introduced by Matheron ~\cite{matheron1975random} and Serra~\cite{serra1982image}. It is found that the medial axis transformation(MAT) or skeleton is one such popular descriptor based on morphology for shape represenation and classification. The skeleton of an object is the locus of the centres of the maximal disks that can be inscribed inside the object boundary. Skeleton is a very useful shape descriptor and is essential for shape representation and analysis in many application areas such as content-based image retrieval systems, character recognition systems, circuit board inspection systems and analysis of biomedical images~\cite{blum1973biological}. Skeletons can capture articulation of shapes in better form than contours~\cite{sebastian2005curves}. 

Skeletons can be organized in the form of attribute-relation graphs (ARG) for the matching purpose. One of the examples for ARG is shock graphs introduced by  Siddiqi et al. ~\cite{siddiqi1999shapesa}~\cite{siddiqi1999shockb}.  Shock Graph is an abstraction of skeleton of a shape onto a Directed Acyclic Graph (DAG). The skeleton  points are labeled according to the radius function at each point. Shock graphs are constructed using the specialized grammar called Shock Grammar ~\cite{siddiqi1996shock}. In the skeleton, the branch points, end-points, and skeleton segments contain both geometrical and topological information. These primitives are referred as shocks~\cite{sebastian2001recognition}~\cite{sebastian2004recognition}. The concept of bone graph is an extension of shock graph. Bone graphs retain the non-ligature structures of the shock graph and are more stable~\cite{macrini2008skeletons}.  There are several algorithms which are proposed in order to have an efficient match of these skeletal graphs. Sebastian et al ~\cite{sebastian2004recognition} proposed a matching approach for node-attributed trees which measures the edit distance between two graphs. Since this method involves complex edit operation, it suffers from high computational complexity.

The techniques which involve graph matching by finding the correspondence between nodes through the conversion of skeleton graphs to skeleton trees require heuristic rules to select the root node~\cite{demirci2006object}~\cite{pelillo1999matching}. The major drawback of this method is that a small change in the shape causes the root to change that results in significant change in the topology of the tree representation. Apart from this, the conversion from graph to a tree structure results in loss of significant structural information and hence leads to wrong match~\cite{bai2008path}. In 2008, Bai and Latecki ~\cite{bai2008path} proposed a method based on the path similarity between the end points of the skeleton graphs. In this method, the geodesic paths between skeleton end points are obtained and are matched using Optimal Subsequence Bijection (OSB) method. Unlike other methods, this method does not convert the skeleton into a graph and matching the graphs,which is still an open problem. This approach addresses the problem of shapes having similar skeleton graphs, but different topological structures and having different skeletons with visually similar shapes. This approach works well even in the presence of articulation and contour deformation.

Demicri et al. ~\cite{demirci2009skeletal} proposed a technique called skeletal shape abstraction from examples. A many to many correspondence between the object parts is used to match the object skeletons. This method is invariant to shape articulation and appears to be more suitable for object detection in edge images. However, the tree abstractions are not specific enough to prevent hallucinating of target object in clutter. The dissimilarity value between the object trees is computed by performing many to many matches between the vertices of the trees. Earth Movers Distance (EMD), under L1 norm, is used as a metric for the matching process. 

Shu and Wu ~\cite{shu2011novel} proposed contour points distribution histogram (CPDH) as the shape descriptor. Though it is not the best one compared to the other state-of-art approaches in terms of retrieval rate, it is relatively simple and has low time complexity. Wang et al. ~\cite{wang2012shape} proposed  a shape descriptor which represents the object contour by a fixed number of sample point. Each sample point is associated with a height function. This method is capable of handling nonlinear deformation of the objects. Shen et al.~\cite{shen2012shape} proposed a method which finds the common structure in a cluster of object skeleton graph. This method outperforms other methods and also suitable for large datasets. However, the time complexity of this method is high due to agglomerative hierarchical clustering.

\section{Morphological Pattern Spectrum Representation}
\begin{figure}[hbtp]
\centering
\includegraphics[scale=0.20]{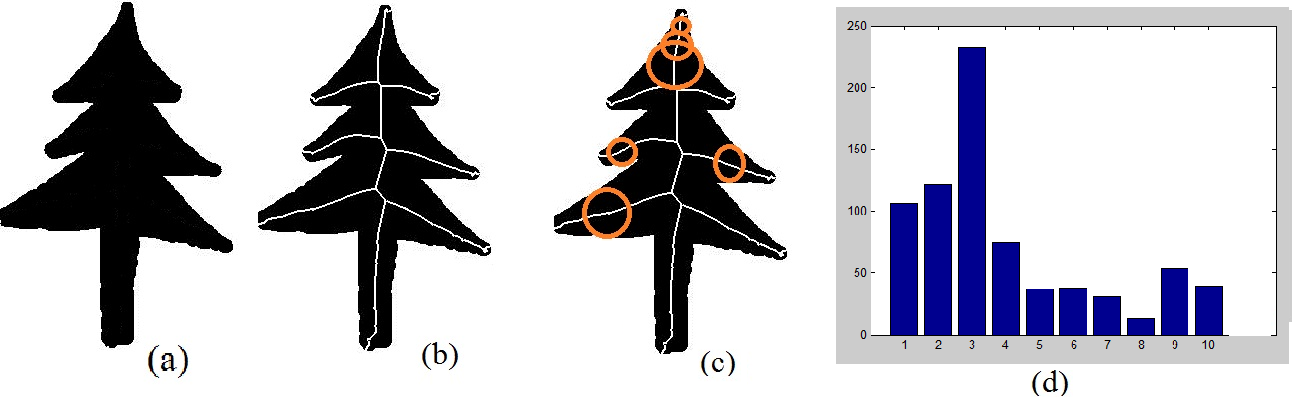}
\caption{(a)Input shape, (b)Skeleton of shape (a), (c) Maximal disc inscribed on the skeleton of the shape(a), (d) Histogram representing the radius of discs and the number of such discs}
\label{fig1}
\end{figure}

The original idea of pattern spectrum proposed by Maragoas ~\cite{maragos1989pattern} is based on Serra's Mathematical morphology filters, that describe the distribution of local figure thickness. The pattern spectrum is a morphological tool that gives the quantitative information about the shape and sizes of the objects in the image. It is invariant to the rotation and shift of the object and can be directly represented using histograms. The size distribution can be represented in the form of histogram for further processing. The additional attributes such as length, orientation and area can be included along with the shape and size. Pattern spectrum focuses on spatial distribution of the skeleton points with respect to the object contour. In order to construct the pattern spectrum, the shape need to be subjected to skeletonization. Then for each skeleton point x, the radius of the maximum circle which can be inscribed inside the shape contour with x  as the centre is determined. The radius function, associated with each skeleton point is  represented in the form of histogram forming the feature vectors.
To illustrate, the skeleton of a shape obtained by applying distance transform method ~\cite{latecki2007skeletonization} and its corresponding thickness map with respect to the object boundary is shown in figure ~\ref{fig1}(b) and (c). The thickness map is later transformed into $10$ bin histogram as shown in figure ~\ref{fig1}(d)

\section{Local Binary Pattern(LBP) Method}
The Local Binary Pattern (LBP) is an excellent metric which focuses on the spatial structure of the image ~\cite{ojala2002multiresolution}. It is a powerful tool for rotation invariant texture classification. The LBP computation on a gray scale image is done as follows.\\
Let $g_{c}$ be the pixel for which the LBP value need to be calculated. Let $g_{p},\ (p=0,\ldots, P-1) $ correspond to the gray values of P equally spaced pixels on a circle of radius $R\left( R > 0 \right)$ that form a circularly symmetric neighbour set of $g_{c}$ . 
% $g_{p}$ ,where $p=0,2, \ldots p-1$ are the gray values of $p$ neighbours around $g_{c}$.  
Without losing information, the gray value of the center pixel $g_{c}$ can be calculated as, 
\begin{equation}
 T =(g_{c}, g_{0}-g_{c}, g_{1}-g_{c}, \ldots, g_{p}-g_{c}). \\
\end{equation}
 
Assuming that  the differences $g_{p}-g_{c}$  are independent of $g_{c}$,  $T$ can be factorized as 
\begin{equation}
T  \approx t(g_{c}) t( g_{0}-g_{c}, g_{1}-g_{c}, \ldots, g_{p-1}-g_{c})  
\end{equation}

The factorized distribution is only an approximation of the joint distribution as exact independence is not there in practice. Hence, though there is a small loss in information existence, this can be tolerable to achieve invariance with respect to shifts in gray scale. The  invariance is achieved with respect to the scaling of the gray scale by considering just the signs of the differences instead of their exact values. 
\begin{equation}
T \approx t( s(g_{0}-g_{c}), s(g_{1}-g_{c}), \ldots,s(g_{p-1}-g_{c})) 
\end{equation}

where  
\begin{eqnarray}
   s(x)  = \left\{ 
  \begin{array}{l l}
    1, &  \ \  if\  x \geq 0  \nonumber    \\
    0, &  \ \ if\   x < 0 
  \end{array} \right\}
\end{eqnarray}

 By assigning a binomial factor $2^{p}$ for each sign $s(g_{p}-g_{c})$, a unique transform $LBP_{P,R}$  is obtained that characterizes the spatial structure of the local image texture:
\begin{equation}
LBP_{P,R}  = \sum_{p=0}^{p-1}  s(g_{p}-g_{c})2^{p}
\end{equation}
  $LBP_{P,R}$ operator is invariant against any monotonic transformation of the gray scale. $LBP_{P,R}$  value remains constant as long as the order of the gray values in the image stays the same.

The $LBP_{P,R}$  operator produces $2P$ different output values, corresponding to the $2^{P}$ different binary patterns that can be formed by the $P$ pixels in the neighbour set. When the image is rotated, the gray values $g_{p}$ will correspondingly move along the perimeter of the circle around $g_{0}$.  Hence, rotating a particular binary pattern naturally results in a different $LBP_{P,R}$ value. To make it rotation invariant, define the local binary pattern as
\begin{equation}
 LBP_{P,R}^{ri} = min\left\lbrace \left( LBP_{P,R},i \right) | i= 0, 1, \ldots P-1 \right\rbrace;
\end{equation}

 where ROR(x, i) performs a circular bit-wise right shift on the P-bit number by i times. In terms of image pixels, (5) simply corresponds to rotating the neighbour set clockwise
so many times that a maximal number of the most significant bits, starting from $g_{p-1}$, is 0. 

In our work, the LBP is computed for the object pixels with 8 neighbours. These LBP values are represented in the form of histogram.

\section{Earth Movers Distance}
The Earth Movers Distance (EMD) is defined as a  minimal cost that must be paid to transform one distribution into the other ~\cite{rubner2000earth}.  It is more robust than histogram matching techniques and can operate on variable-length representations of the distribution. For instance, if Ha and Hb are two histograms, EMD is the minimum amount of work needed to transform histogram Ha towards Hb. Given two distributions, it can be seen as a mass of earth properly spread in space and the other as a collection of holes in the same space. Then, the EMD measures the least amount of work needed to fill the holes with earth. 

To illustrate, let P = ${(p_{1},wp_{1}),...,(p_{m},wp_{m})}$  be the first distribution  with $m$ clusters, where $p_{i}$ is the cluster representative and $wp_{i}$ is the weight of the cluster and Q = ${(q_{1},wq_{1}),...,(q_{m},wq_{m})}$ be the second distribution with $n$ clusters. In other words, $p_{i}$and $q_{i}$ typically represent bins in a fixed partitioning of the relevant region of the underlying feature space, and the associated $wp_{i}$ and $wq_{i}$ are a measure of the mass of the distribution that falls into the corresponding bin. Let D= $d_{ij}$ be the ground distance matrix where $d_{ij}$ is the ground distance between clusters $p_{i}$ and $q_{j}$. We need to estimate the flow F = $[ f_{ij}]$, with $f_{ij}$ be the flow between $p_{i}$ and $q_{j}$ , that minimizes the overall cost. The Minimal cost (WORK), between two distribution is calculated as follows:
\begin{eqnarray}
WORK(P,Q,F) = \sum_{i=1}^{m}\sum_{j=1}^{n} d_{i j} f_{i j}, \ \ s.t \ \  f_{i j} \geq 0 \nonumber
\end{eqnarray}
Subject to the following constraints
\begin{eqnarray}
\sum_{j=1}^{n} f_{i j} \leq W_{pi},  \  &1& \leq i \leq m  \ ;\ \ \
\sum_{i=1}^{m} f_{i j}\leq W_{qi},  \ \   1 \leq j \leq m    \nonumber  \\
\sum_{i=1}^{m}\sum_{j=1}^{n} f_{i j} &=& min\left(\sum_{i=1}^{m} W_{pi}, \sum _{j=1}^{n} W_{qj}\right)  
\end{eqnarray}

Constraint(1) specifies that the supplies can be moved from P to Q and not vice versa. Constraint(2) limits the amount of supplies that can be sent by the clusters in P to their weights. Constraint(3) limits the clusters in Q to receive no more supplies than their weights and constraint(4) forces to move the maximum amount of supplies possible. The amount of flow is called as the total flow. Once the transportation space problem is solved, and the optimal flow F is found, the earth movers distance is defined as the resulting work normalized by the total flow:
\begin{equation}
EMD(P,Q) = \frac{\sum_{i=1}^{m} \sum_{j=1}^{n} d_{i j} f_{i j}}{\sum_{i=1}^{m}\sum_{j=1}^{n} f_{i j}}
\end{equation}

\section{Proposed approach}
In the proposed work, a combined approach of morphological pattern spectrum and local binary pattern are used to extract the shape features. The LBP of the shape is computed. Both pattern spectrum and LBP features are represented in the form of histogram forming the hybrid pattern spectrum knowledge base. The EMD distance measure with L1 ground distance is used for comparing the features represented in the form of histogram. 

\subsection{Feature extraction by Morphological Pattern spectrum} 
The morphological pattern spectrum is computed for the skeleton of the shape as well as the contour. The steps involved in this process are given below
\begin{enumerate}

\item The skeleton of the shape is extracted using skeleton strength map(SSM)~\cite{latecki2007skeletonization} which is based on the Euclidean distance transform method.
\item The pattern spectrum need to be computed for each skeleton point. The radius function of each skeleton point, with respect to the shape boundary is obtained. The radius function value for a pixel x is the radius of the maximal disc inscribed inside the object with x as a skeleton point. The radius value for all the skeleton points forms the pattern spectrum of the shape. 
\item To obtain the radius values for the skeleton points, skeleton strength map(SSM)~\cite{latecki2007skeletonization} based on the Euclidean distance transform is used. The forward mask shown in Figure~\ref{fig2}(a) and 90 degree rotated backward mask given in Figure~\ref{fig2}(b) are used as structuring elements. The forward mask is applied linearly over the block, which is of same size as the shape, in the forward direction i.e., from left to right and top to bottom, resulting in forward pass. The backward mask is applied over the image in reverse direction from right to left and bottom to top resulting in backward pass. The block is aligned with the shape skeleton and the values in the distance map, corresponding to the skeleton point, represents the radius. The  result of forward pass and reverse pass of the block and the equivalent distance maps are depicted in figure ~\ref{fig3}(d)-(f).
\begin{figure}[hbtp]
\centering
\includegraphics[scale=0.50]{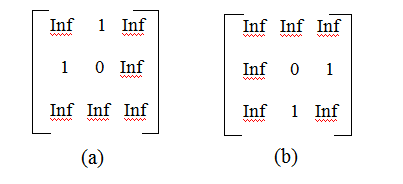}
\caption{ (a) Forward mask (b) Backward Mask}
\label{fig2}
\end{figure}

\begin{figure}[hbtp]
\centering
\includegraphics[scale=0.350]{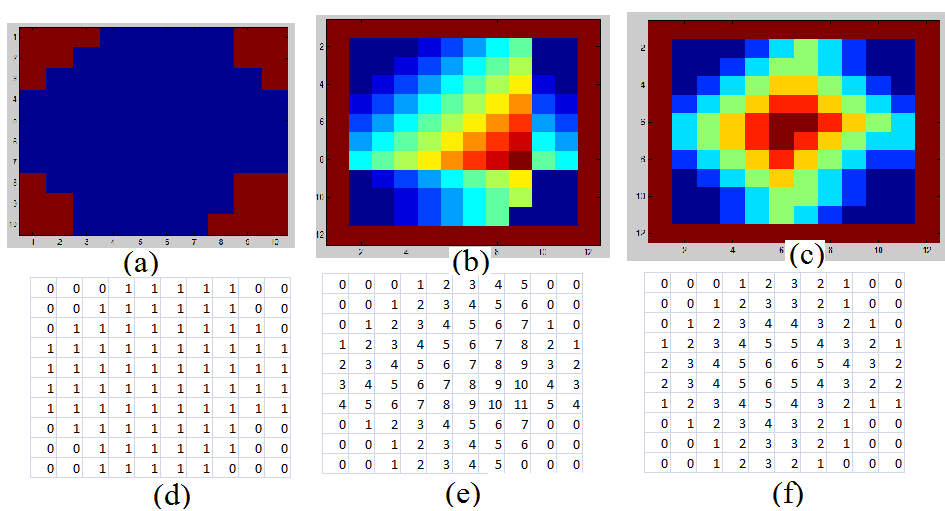}
\caption{Shape B (b) Result of the forward pass on B (c) Result of the backward pass
on B , (d), (e), (f) represents the corresponding distance maps of (a), (b) and (c)}
\label{fig3}
\end{figure}

\item The pattern spectrum thus obtained is converted to a histogram with 10 bins represents shape feature.
\item The above process is repeated for all the shapes in the training set to form the Pattern spectrum knowledge base of skeleton points.
\item A bounded box is fitted to the shape to eliminate the non shape portion of the image and the contour of the shape is extracted
\item  For each contour pixel, the pattern spectrum is computed with the bounding box as the reference line. The radius function is calculated using a 3 x 3 structuring element. Results of forward and backward mask is shown in figure ~\ref{fig4}
\item The radius function values obtained for each contour pixel is represented in the form of histogram forms the pattern spectrum knowledge base of contour points. 
\end{enumerate}

\begin{figure}[hbtp]
\centering
\includegraphics[scale=0.20]{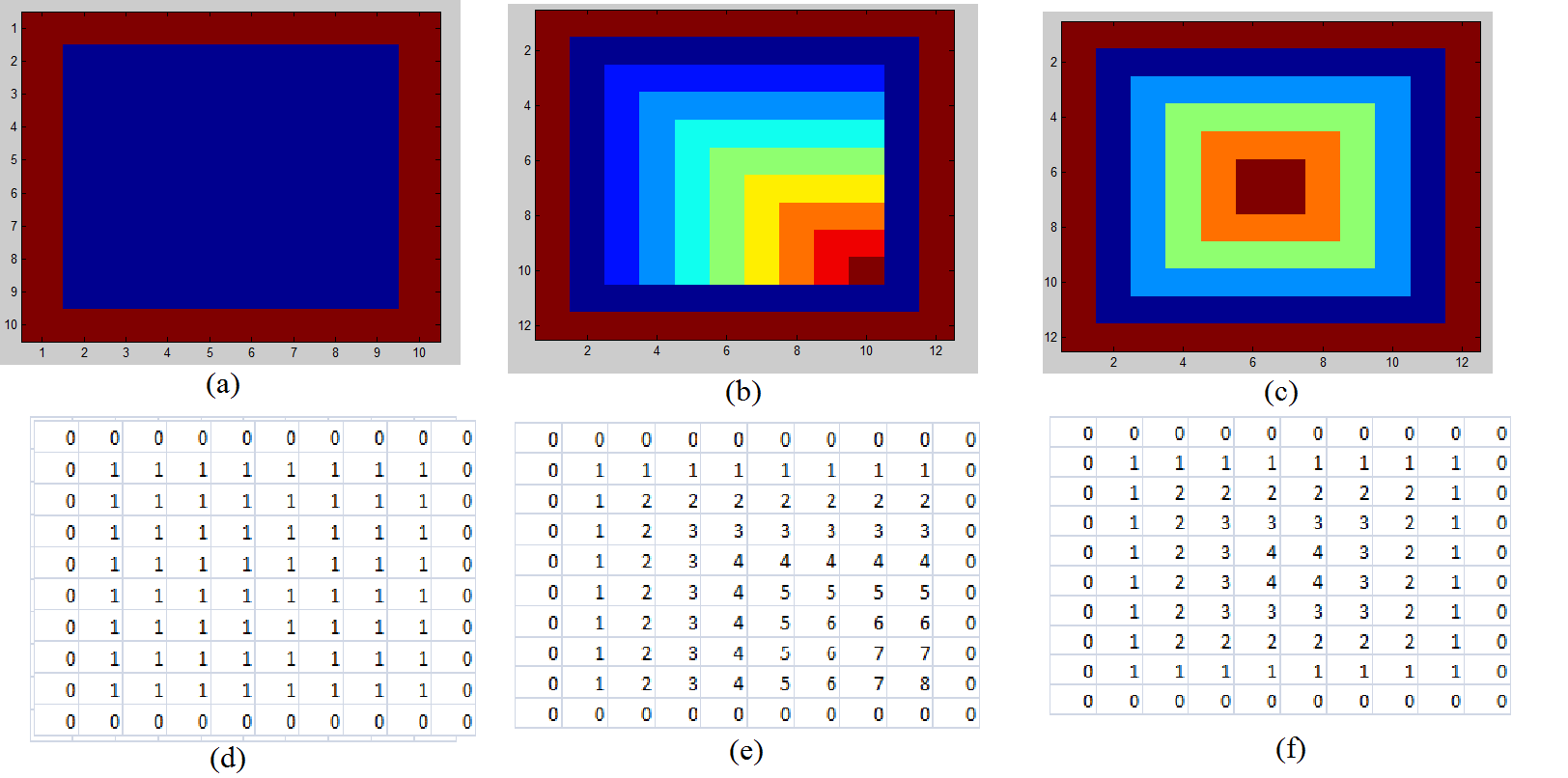}
\caption{a) Block B (b) Result of the forward pass on B (c) Result of the backward pass
on B , (d), (e), (f) represents the corresponding distance maps of (a), (b) and (c)}
\label{fig4}
\end{figure}

\subsection{Feature extraction by LBP method}

The LBP value is computed for each shape pixel as follows.
For each pixel, identify the 8 neighbours of the pixels. Obtain the LBP sequence of 8 neighbours. Compute the decimal equivalence of the this binary string. The string is subjected to circular right shift by 1, and then its decimal value is calculated. This is repeated for 8 times obtaining 8 decimal values. The binary string corresponding to smallest decimal value is taken forming the LBP value of the pixel  under processing.
The array of LBP values for each shape pixels is now represented in the form of histogram. This process is repeated for each shape in the training set forming the LBP knowledge base of the training set. Thus every shape is described by three histograms obtained due to Skeleton, Contour spectrum and LBP and thus forms a fused feature for every shape.

%The steps involved in the proposed approach for off-line signature verification is as shown in figure~\ref{fig2}.
%
%\begin{figure}[hbtp]
%\centering
%\includegraphics[scale=0.75]{pattern3.png}
%\caption{Block diagram of off-line signature verification}
%\label{fig2}
%\end{figure}

\subsection{Classification}
 The Earth movers distance (EMD) is used as the distance measure for comparing the shape features which are represented as histogram. Given a shape feature for its classification, the match is obtained by EMD distance as discussed in section V. 
 
Given the query shape, the classification is done as follows. 

\begin{enumerate}
\item Extract the skeleton pattern spectra (SPS) of the test shape and obtain its histogram, say St
\item Extract the contour pattern spectra(CPS) of the test shape and obtain its histogram, say Ct
\item Compute the LBP of the test shape (LBS) and obtain its histogram, say Lt.
%\item For each shape feature represented in the form of histogram in the skeleton pattern spectrum knowledge base, compute the EMD distance with SPS. This will give the distance matrix say $DS$. 
%For each shape feature represented in the form of histogram in the contour pattern spectrum knowledge base compute the EMD distance with CPS. This results in a distance matrix say $DC$.
%For each shape feature represented in the form of histogram in the LBP pattern spectrum knowledge base compute the EMD distance with LBS. This results in a distance matrix say $DL$.
\item Compute the distance between the test shape with the sample shapes of the training set. Let $D_{S}$, $D_{C}$ and $D_{L}$ be the distances obtained due to EMD distance measure.
\item  The distance matrices $DS$, $DC$ and $DL$ are fused to form the resultant matrix $DR$ as follows
\begin{equation}
DR = \alpha DS + \beta DC + \gamma DL 
\end{equation}
where the values of $\alpha$, $\beta$ and $\gamma$ are deducted experimentally, which are specific to the dataset.
\item  The EMD distance values are arranged in the ascending order of the distance to obtain top shape matches. 
  
\end{enumerate}

\section{Experimental Results and Discussions}
In this section, we present the experimental results conducted on the standard shape datasets namely: \textit{Kimia-99}, \textit{Kimia-216} and \textit{MPEG-7}. The performance of the proposed approach is demonstrated through \textit{Bull's eye score}~\cite{ling2007shape}. All experiments are conducted using MATLAB tool and tested on Pentium(R) dual core CPU with 3GB RAM on Windows-7. 

\subsection{Experimental results on Kimia-99}
\begin{figure}[hbtp]
\centering
\includegraphics[scale=0.4]{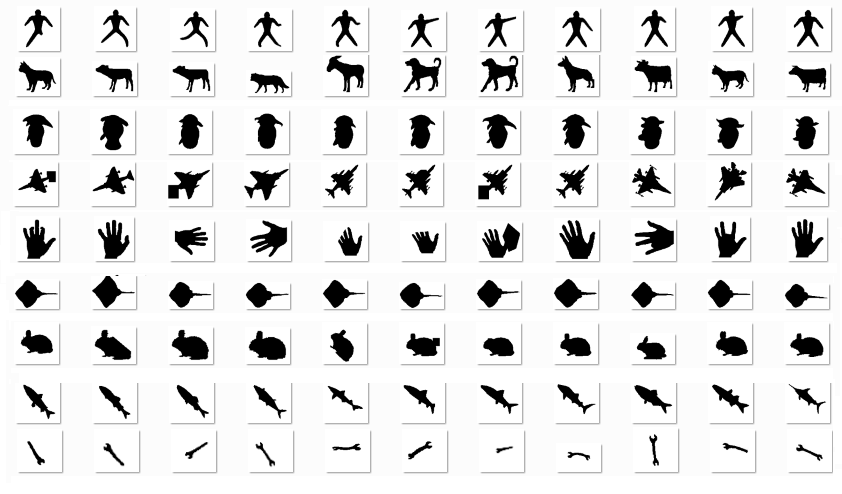}
\caption{Kimia 99 dataset: 9 classes with 11 samples in each class}
\label{fig5}
\end{figure}

Kimia-99 shape dataset consists of 9 classes with 11 samples in each class. The 99 shapes belong to 9 different classes are shown in Figure~\ref{fig5}. The retrieval rate depicting the top 10 matching shapes is presented in Table~\ref{tab1}. We have also given the retrieval results obtained due to  Belongie et al.~\cite{belongie2002shape}, Shu and Wu~\cite{shu2011novel} and Tu and Yuille~\cite{tu2004shape}. It shall be observed from Table~\ref{tab1} that the proposed approach exhibits better performance when compared to other approaches.

%the  gives the 99 shapes of 9 classes from Kimia-99 dataset. The coefficients $\alpha$, $\beta$ and $\gamma $ used for fusing the distance matrices $DS$, $DC$ and $DL$ possess the values $0.7$, $0.3$ and $0.1$ respectively. From the literature we observed that Belongie et al.~\cite{belongie2002shape}, Shu and Wu~\cite{shu2011novel} and Tu and Yuille~\cite{tu2004shape} have experimented on Kimia-99 dataset and hence a comparative analysis along with the performance of the proposed approach is summarized in Table~\ref{tab1}. The table also depicts the number of top 1 to top 10 closest matches of the query shape to the retrieved shape class.  The best possible result for each category is $99$.

\begin{table}
\centering
\caption{Top 10 closest matching shapes}
\begin{tabular}{|p{1.15cm}|p{0.2cm}|p{0.2cm}|p{0.2cm}|p{0.2cm}|p{0.2cm}|p{0.2cm}|p{0.2cm}|p{0.2cm}|p{0.2cm}|p{0.2cm}|p{0.2cm}|}
\hline 
Approach &1st &2nd &3rd &4th& 5th& 6th &7th& 8th& 9th& 10th& total \\
\hline
SC ~\cite{belongie2002shape} & 97 &  91 &  88 &  85 &  84 &  77 &   75 &   66 &  56 &  37 &   756  \\ 
\hline 
CPDH+EMD (Eucl)~\cite{shu2011novel} & 96&   94&   94&   87&   88&   82&   80&   70&   62&   55 & 808 \\ 
\hline
CPDH+EMD (shift)~\cite{shu2011novel} &98 &  94&   95 &  92 &  90 &  88 &  85 &  84 &  71 &  52  &  849 \\ 
\hline 
Gen. model ~\cite{tu2004shape} &99 &  97 &  99 &  98 &  96 &  96 &  94 &  83 &  75 &  48  &  885 \\ 
\hline 
\textbf{Proposed Approach}  &99 &  97 &  97&   88&   88&   86&   86&   90&   80&   77   & 888 \\ 
\hline 
\end{tabular}
\label{tab1} 
\end{table}

\subsection{Experimental results on Kimia-216}

\begin{figure}[hbtp]
\centering
\includegraphics[scale=0.50]{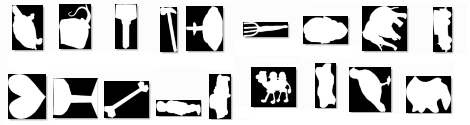}
\caption{Kimia 216 dataset : A sample shape from each class}
\label{fig6}
\end{figure}

Kimia-216 shape dataset consists of 18 classes with 12 samples in each class. One such sample from each class is shown in Figure~\ref{fig6}. The top 11 closest matches obtained due to the proposed methodology is shown in Table~\ref{tab2}. In Table~\ref{tab2}, we have also given the retrieval results obtained due to Belongie et al.~\cite{belongie2002shape}and Shu and Wu~\cite{shu2011novel}. One can notice here too that the proposed approach works on par with the existing methods.    

 %Figure~\ref{fig6} gives one sample shape from each class of Kimia-216 dataset.   The coefficients $\alpha$, $\beta$ and $\gamma $ used for fusing the distance matrices $DS$, $DC$ and $DL$ possess the values $3.9$, $16.0$ and $0.1$ respectively. From the literature we observed that Belongie et al.~\cite{belongie2002shape}and Shu and Wu~\cite{shu2011novel} have experimented on Kimia-216 dataset and hence a comparative analysis along with the performance of the proposed approach is summarized in Table~\ref{tab2}. The table also depicts the number of top 1 to top 11 closest matches of the query shape to the retrieved shape class.  The best possible result for each category is $216$.

\begin{table}
\centering
\caption{Top 11 closest matching shapes}
\begin{tabular}{|p{1.05cm}|p{0.15cm}|p{0.15cm}|p{0.15cm}|p{0.15cm}|p{0.15cm}|p{0.15cm}|p{0.15cm}|p{0.15cm}|p{0.18cm}|p{0.18cm}|p{0.18cm}|p{0.3cm}|}
\hline 
Approach &  1st &2nd &3rd &4th& 5th& 6th &7th& 8th& 9th& 10th& 11th &Total  \\
\hline
SC ~\cite{belongie2002shape} & 214 & 209 & 205&  197 & 191 & 178 & 161 & 144&  131 & 101  & 78 & 1809  \\ 
\hline 
CPDH+ EMD (Eucl)~\cite{shu2011novel} & 214 & 215 & 209&  204&  200&  193&  187&  180&  168&  146 & 114 & 2030\\
\hline
CPDH+ EMD (shift)~\cite{shu2011novel} &215 & 215&  213&  205&  203 & 204&  190&  180&  168&  154&  123 & 2070 \\ 
\hline 
\textbf{Proposed Approach}& 216& 209&  205&  195&  195& 197& 188& 180& 179& 163& 152 & 2079 \\ 
\hline 

\end{tabular}
\label{tab2} 
\end{table}

\subsection{Experimental results on MPEG-7 :}
\begin{figure}[hbtp]
\centering
\includegraphics[scale=0.350]{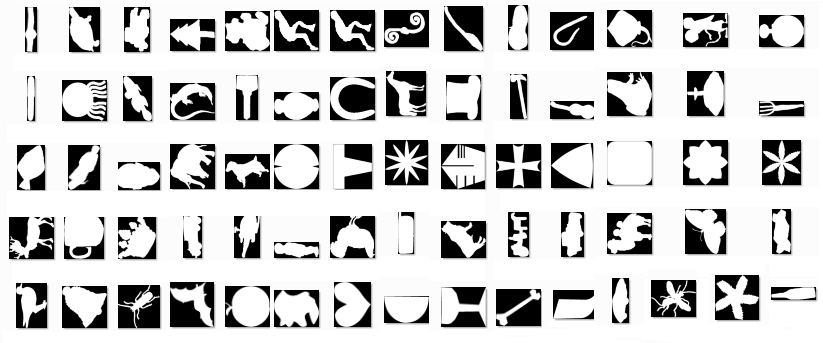}
\caption{MPEG-7 dataset: A sample shape from each class}
\label{fig7}
\end{figure}

\begin{figure*}[hbtp]
\centering
\includegraphics[scale=0.55]{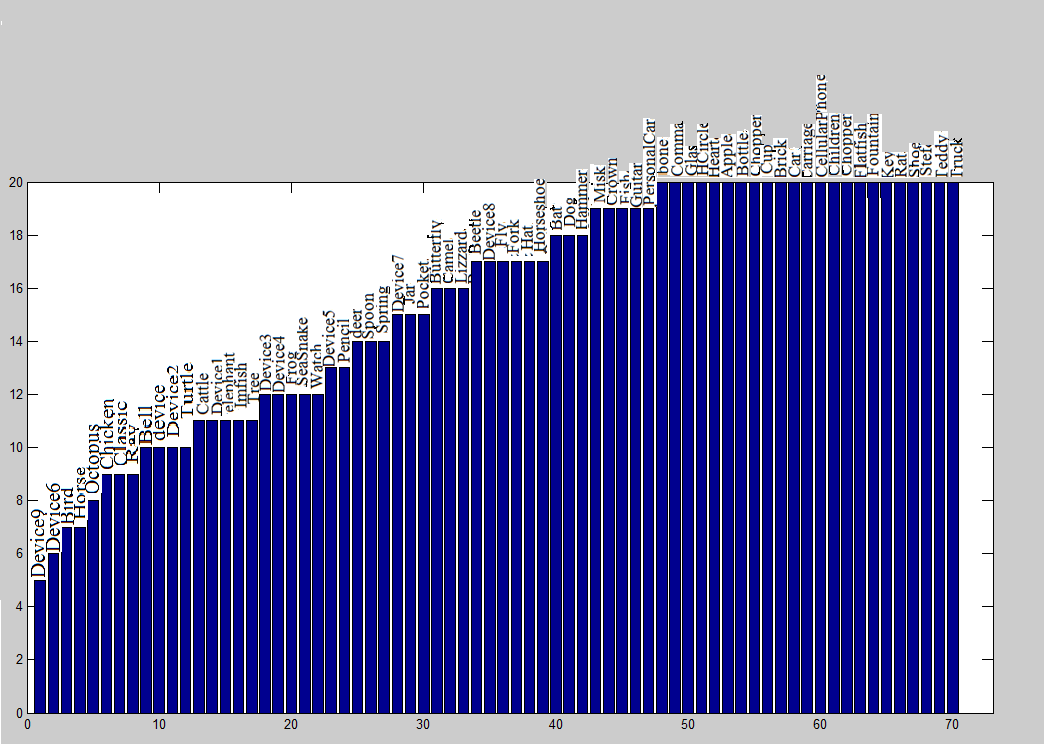}
\caption{Clas-wise retrieval results for MPEG-7 dataset}
\label{fig8}
\end{figure*}

MPEG-7 dataset consists of 1400 shapes from 70 classes with each class consisting of 20 shape samples. Figure~\ref{fig7} shows one sample shape from each class of MPEG-7. The dataset is challenging due to the presence of shapes that are visually dissimilar from other members of their class and shapes that are highly similar to members of other classes. The retrieval rate of the proposed approach on MPEG-7 is calculated and the results obtained due to proposed approach along with the results of the state-of-art algorithms is tabulated in Table~\ref{tab6}. The retrieval rate obtained due to proposed approach for each shape class considering top 40 retrievals is shown in Figure~\ref{fig8} and also tabulated in Table~\ref{tab3}. The retrieval rate depicting top 12 closest matching shape is given Table~\ref{tab4}.

% state using bull's eye score, along with the comparative analysis with the state-of-art algorithms is tabulated in Table ~\ref{tab4}. The bins representing average scores per shape class considering top 40 retrievals on MPEG-7 is shown in figure ~\ref{fig8}. The name of the class and values corresponding to each bin in the fig  ~\ref{fig8} is given in the table ~\ref{tab3}. The number of top 1 to top 12 closest matches of the query shape from MPEG-7 is tabulated in Table~\ref{tab6}.  The best possible result for each category is $1400$.

Table ~\ref{tab5} provides overall performance of the proposed approach measured using bull's eye score on Kimia-99, Kimia-216 and MPEG-7 shape datasets.

\begin{table*}
\centering
\caption{Clas-wise retrieval results for MPEG-7 dataset}
%\begin{tabular}{|p{1cm}|p{0.28cm}|p{0.3cm}|p{0.28cm}|p{0.3cm}|p{0.28cm}|p{0.28cm}|p{0.25cm}|p{0.25cm}|p{0.25cm}|p{0.4cm}|}
%\begin{tabular}{|c|p{0.7cm}|p{0.7cm}|p{0.7cm}|p{0.7cm}|p{0.7cm}|p{0.7cm}|p{0.7cm}|p{0.5cm}|p{0.5cm}|p{0.5cm}|}
\begin{tabular}{|c|c|c|c|c|c|c|c|c|c|c|}
%\hline 
%Class & 1st &2nd &3rd &4th& 5th& 6th &7th& 8th& 9th& 10th\\
\hline 
Class &Device9 &Device6 &Bird& Horse& Octopus& Chicken& Classic& Ray& Bell& device\\
\hline
MPEG-7 &5 &	6&	7&	7&	8&	9	&9	&9	&10&	10 \\
%\hline
%Class  & 11st &12nd &13rd &14th& 15th& 16th &17th& 18th& 19th& 20th \\
\hline
Class  & Device2 &Turtle& Cattle &Device1 &elephant &Imfish& Tree& Device3 &Device4& Frog \\
\hline
 MPEG-7 &10	&10	&11&	11&	11&	11&	11&	12	&12&	12\\	
\hline
 %Class & 21st &22nd &23rd &24th& 25th& 26th &27th& 28th& 29th& 30th \\	
%\hline
 Class &SeaSnake& Watch &Device5& Pencil& deer& Spoon& Spring& Device7 & Jar& Pocket, \\	
\hline
 MPEG-7 &12&	12	&13	&13	&14	&14	&14	&15&	15	&15\\	
%\hline
% Class & 31st &32nd &33rd &34th& 35th& 36th &37th& 38th& 39th& 40th\\	
\hline
 Class & Butterfly& Camel& Lizzard& Beetle& Device8 &Fly&Fork& Hat&	Horseshoe& Bat\\
 \hline
 MPEG-7 &16	&16&	16&	17&	17&	17&	17&	17&	17&	18\\	
\hline
 %Class & 41st &42nd &43rd &44th& 45th& 46th &47th& 48th& 49th& 50th\\	
%\hline
Class & Dog& Hammer& Misk& Crown& Fish& Guitar& PersonalCar&	bone&   Comma&     Glas\\	
\hline
 MPEG-7 &18	&18&	19&	19&	19&	19&	19&	20&	20&	20\\	
\hline
 % Class & 51st &52nd &53rd &54th& 55th& 56th &57th& 58th& 59th& 60th \\
%\hline
  Class & HCircle& Heart& Apple& Bottle& Chopper& Cup&Brick &Car& Carriage& CellularPhone  \\
\hline
 MPEG-7 &20&	20&	20&	20&	20&	20&	20&	20&	20&	20\\
\hline
 % Class & 61st &62nd &63rd &64th& 65th& 66th &67th& 68th& 69th& 70th \\
%\hline
  Class &Children& Chopper&Flatfish&  Fountain &Key& Rat& Shoe & Stef& Teddy& Truck \\
\hline
 MPEG-7 &20&	20&	20&	20&	20&	20&	20&	20&	20&	20\\	
\hline
\end{tabular}
\label{tab3} 
\end{table*}

%bone, Comma, Glas, HCircle, Heart, Misk,Apple, Bat, Beetle, Bell
%Bird,Bottle,Brick,Butterfly,Camel,Car, Carriage,Cattle,CellularPhone,Chicken
%Children,Chopper,Classic,Crown,Cup,deer,device ,Device1,Device2, Device3
%Device4, Device5,Device6, Device7,Device8,Device9,Dog,elephant,Face,Fish
%Flatfish,Fly,Fork,Fountain,Frog,Guitar,Hammer,Hat,Horse,Horseshoe,
%Jar,Key,Lizzard,Imfish,Octopus,Pencil,PersonalCar,Pocket,Rat,Ray
%SeaSnake, Shoe,Spoon,Spring,Stef, Teddy,Tree ,Truck,Turtle Watch

\begin{table}
\centering
\caption{Retrieval rate(Bulls eye Score) of MPEG-7 dataset- A comparative Analysis}
\begin{tabular}{|c|c|}
\hline 
DataSet & MPEG-7  \\ 
\hline
CSS ~\cite{mokhtarian1997efficient}  &75.44 \\
\hline
Visual parts~\cite{latecki2000shape}& 76.45\\
\hline
SC ~\cite{belongie2002shape} &  76.51 \\ 
\hline 
CPDH ~\cite{shu2011novel} & 76.56  \\
\hline
Aligning curves ~\cite{sebastian2003aligning} & 78.16 \\
\hline
\textbf{Proposed Approach }& 79.38 \\
\hline
SSC ~\cite{xie2008shape} & 79.92\\
\hline 
\end{tabular}
\label{tab6} 
\end{table}

%  SC  belongie2002shape , IDSC  ling2007shape,  generative model tu2004shape,  SSC xie2008shape,CSS mokhtarian1997efficient
%CPDH

 \begin{table}
\centering
\caption{Bull Eye Score(Retrieval Rate) of proposed method for Kimia 99, Kimia 216, MPEG-7 dataset}
\begin{tabular}{|c|c|}
\hline 
DataSet & Bull eye test  \\ 
\hline
Kimia-99&  99.08 \\ 
\hline 
Kimia-216 & 95.25 \\ 
\hline
MPEG-7 &  79.38\\ 
\hline 
\end{tabular}
\label{tab5} 
\end{table}

\begin{table}
\centering
\caption{Top 12 closest matching shapes}
\begin{tabular}{|p{0.5cm}|p{0.28cm}|p{0.28cm}|p{0.28cm}|p{0.28cm}|p{0.28cm}|p{0.28cm}|p{0.25cm}|p{0.25cm}|p{0.25cm}|p{0.2cm}|p{0.2cm}|p{0.2cm}|}
\hline 
Dataset & 1st &2nd &3rd &4th& 5th& 6th &7th& 8th& 9th& 10th& 11th & 12th\\
\hline 
 MPEG -7 &1400&	1345&	1276&	1221&	1157&	1113&	1070&	1023&	995&	961&	933	&898 \\ 
\hline
\end{tabular}
\label{tab4} 
\end{table}

\section{Conclusion}
In this paper, we explored the application of morphological pattern spectra and Local binary pattern method on binary shapes to extract the dominant features based on pattern spectra followed by classification using EMD. The Pattern spectra is based on the concept introduced by Maragos transformation(Thickness map). The relationship between the Maragos morphological pattern spectrum and the histogram is well utilized here. Also the fusion of LBP method along with the pattern spectra gives better results at a comparable time complexity. It is proved that the morphological spectra and LBP are invariant to rotation and shift of the shape. This method is proven to be computationally efficient. It is experimentally shown that morphological spectra together with LBP, in combination with EMD metric, with L1 as the ground distance can be successfully used in the shape based object classification. The proposed approach is relatively simple to implement, and accurate in terms of shape representation and classification.

\bibliographystyle{IEEEtranS}
\bibliography{Ref_PS}

\end{document}